\documentclass[twocolumn,letterpaper,10pt]{article}
\usepackage{kuz}
\usepackage{graphicx}       				
\usepackage[utf8]{inputenc}  				
\usepackage[round,authoryear]{natbib}
\usepackage[labelfont=bf]{caption}			
\usepackage[colorlinks=false]{hyperref} 	
\usepackage{amsmath}
\usepackage{xcolor}
\usepackage{booktabs}

\usepackage{lipsum}

\makeatletter
\def\convertto#1#2{\strip@pt\dimexpr #2*65536/\number\dimexpr 1#1}
\makeatother

\begin{document}

\date{}
\title{How Should We Teach Robots? A Comparison of Kinesthetic, Joystick, and Gesture-Based Teaching}

\author{
\vspace{1ex}
\textbf{Petr Vanc,\quad Jan Kristof Behrens,\quad Václav Hlaváč, and\quad Karla Stepanova} \\ 
Czech Institute of Informatics, Robotics and Cybernetics (CIIRC CTU)\\
 Jugoslávských partyzánů 1580/3, 160 00 Dejvice\\
Email: petr.vanc@cvut.cz \\
}

\maketitle 

\thispagestyle{empty}

\noindent
{ \begin{center} \bf\normalsize Abstract\end{center}}
{
\noindent
Instructing robots from demonstrations can be done through different teaching modalities, each with different usability and performance trade-offs. This paper compares kinesthetic guidance, joystick teleoperation, and hand gestures in a user study with eight participants. We evaluate replay success, modified NASA-TLX workload, and common teaching errors across three manipulation tasks. Kinesthetic guidance produced the shortest demonstrations, lowest workload, and highest success on the more orientation-sensitive and contact-rich tasks. Joystick teleoperation performed best on simple peg picking. Hand-gesture teaching, although less reliable overall, performed better than expected and in some cases achieved results comparable to kinesthetic guidance.
}

\section{Introduction}
\label{sec:intro}

Robots are still too often programmed by specialists, even when the desired task could be shown more easily than described in code. Learning from Demonstration (LfD) and Programming by Demonstration (PbD) address this problem by allowing users to teach robots through examples rather than manually specifying trajectories or control policies~\citep{argall2009survey,billard2016learning,ravichandar2020recent}. For non-expert users, however, the question is not only \emph{what} the robot should learn, but also \emph{how} the user should demonstrate it.

The teaching modality strongly shapes the resulting demonstration. Kinesthetic teaching, where the user physically guides the robot arm, is often treated as the most natural option because it gives direct control over the end-effector. Yet it requires safe access to the robot, a compliant control mode, and a user who can comfortably move the robot. These assumptions may fail in remote operation, safety-restricted workspaces, limited-access setups, or accessibility-sensitive scenarios. In such cases, joystick teleoperation or contact-free hand teleoperation may be more practical, although they introduce their own limitations: stepwise control through discrete inputs, hand tracking noise, and difficulty with fine orientation.

\begin{figure}
    \centering
    \includegraphics[width=0.32\linewidth]{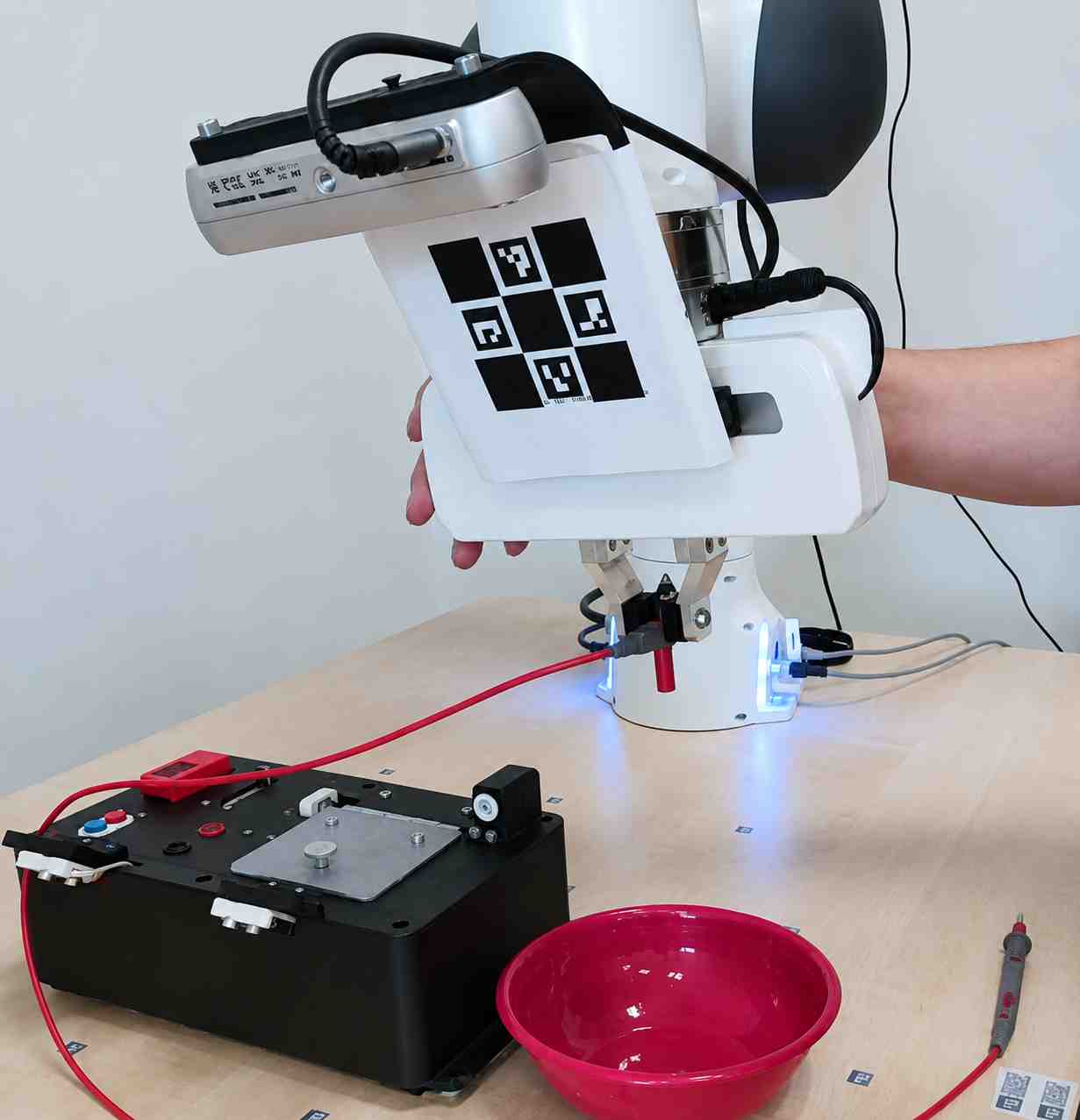}
    \includegraphics[width=0.32\linewidth]{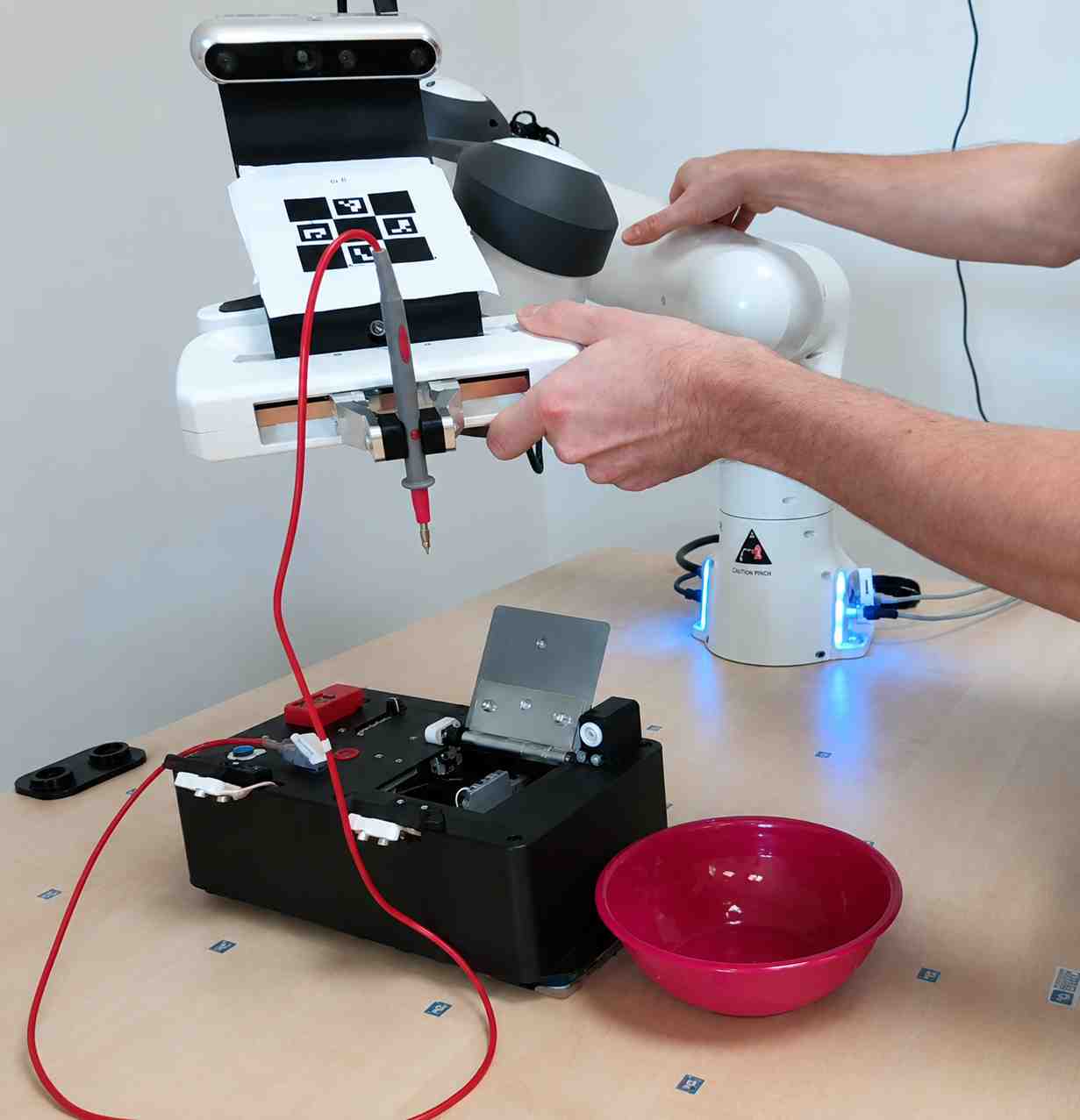}
    \includegraphics[width=0.32\linewidth]{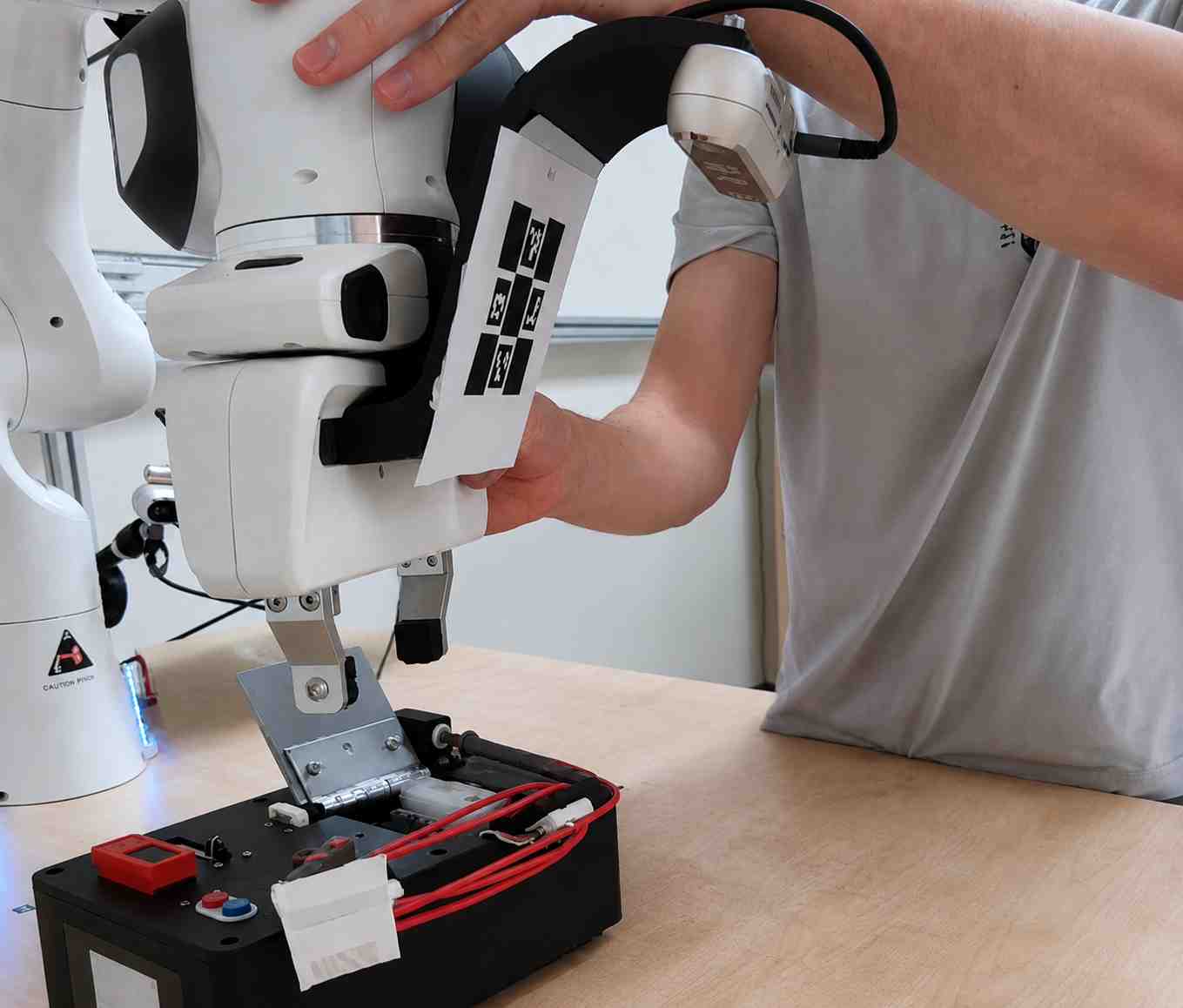}
    \includegraphics[width=0.32\linewidth]{fig/demo02.jpg}
    \includegraphics[width=0.32\linewidth]{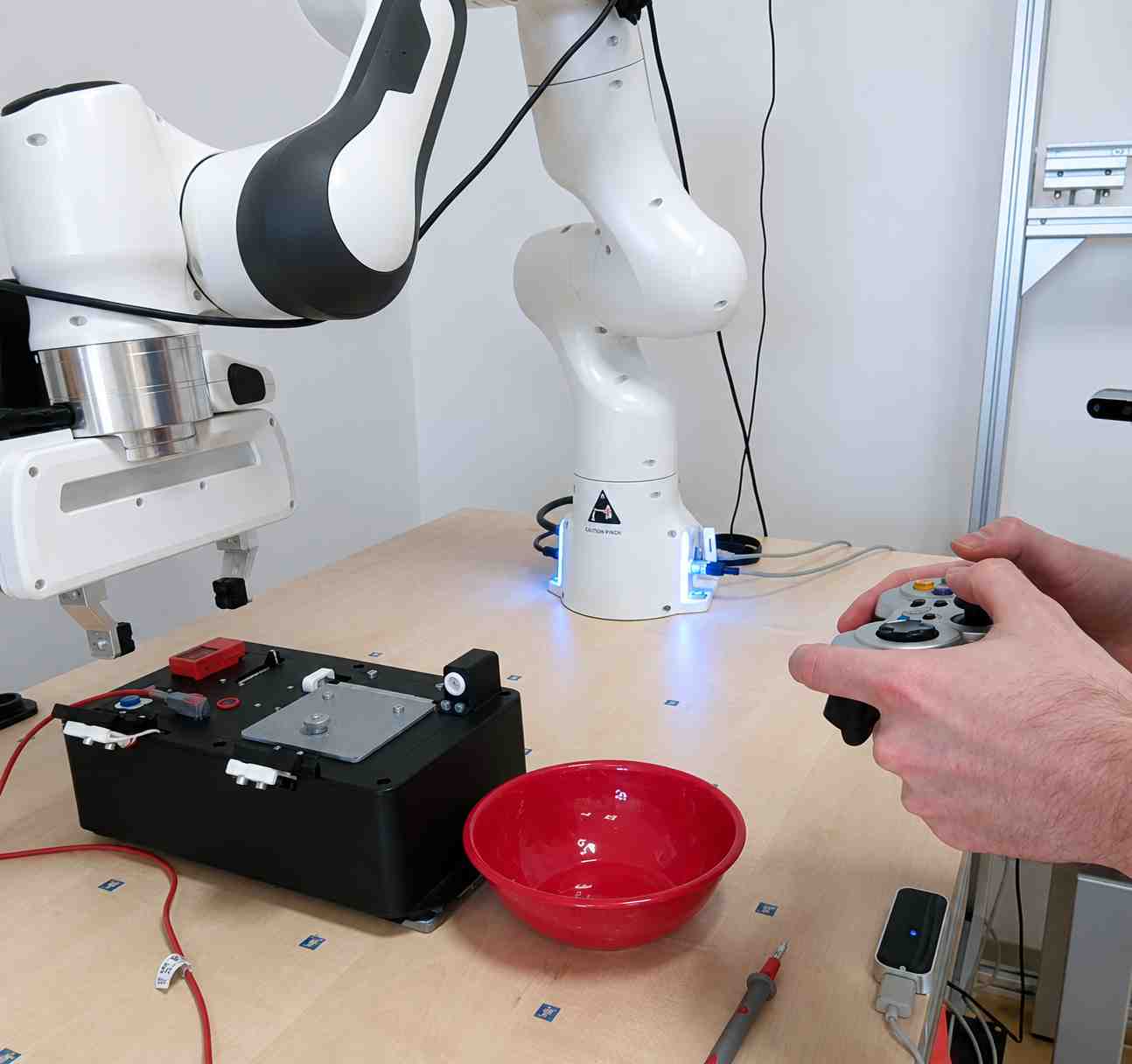}
    \includegraphics[width=0.32\linewidth]{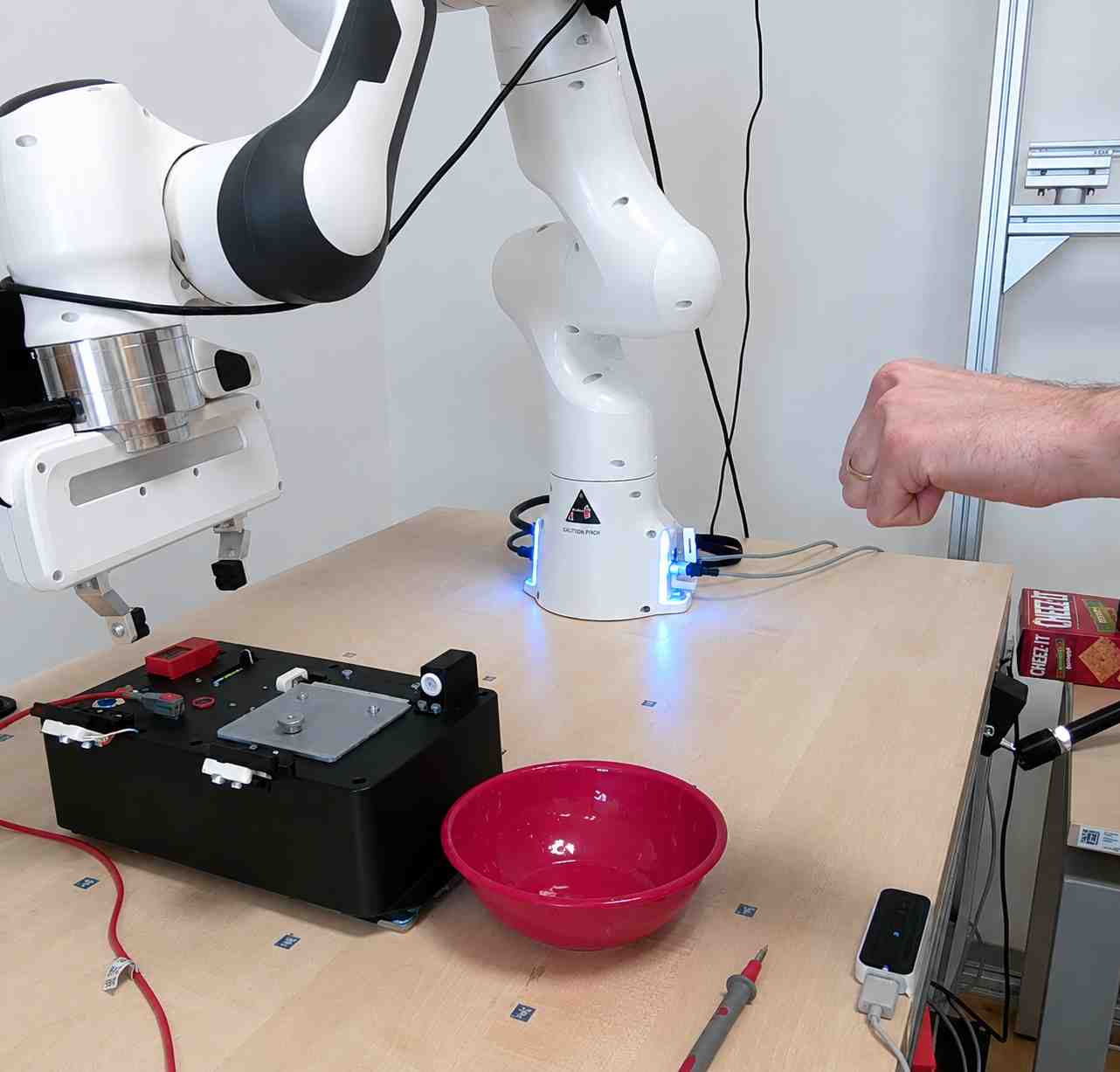}
    \caption{Teaching robot manipulation tasks on the Robothon taskboard. The three evaluated tasks are shown in the top row: (left) peg picking, (center) probe measurement, and (right) cable wrapping. The bottom row visualizes the three teaching modalities used in the study: (left) kinesthetic teaching, (center) joystick teleoperation, and (right) direct teleoperation using hand gestures.}
    \label{fig:intro}
\end{figure}

This paper compares three teaching modalities for teaching robots from demonstrations: kinesthetic guidance, joystick teleoperation, and hand gestures. We analyze a dataset in which eight participants taught three manipulation tasks under multiple scene variants. 
Each recorded variant is treated as a demonstration instance. We evaluate demonstration acquisition and replay quality, not learned policy generalization: replay success measures whether the robot can reproduce the recorded behavior under the corresponding demonstrated task condition. This allows us to compare replay success, demonstration duration, user workload measured by a modified NASA-TLX questionnaire, user preference, and common teaching errors.

We address the following research questions:
\begin{itemize}
    \item How do kinesthetic guidance, joystick teleoperation, and hand gestures compare in replay success and demonstration duration across manipulation tasks?
    \item What workload trade-offs do these teaching modalities impose on users, especially in terms of mental demand, physical demand, effort, and frustration?
    \item Can hand-gesture teaching serve as a viable contact-free alternative when kinesthetic teaching is unavailable, unsafe, or physically demanding?
\end{itemize}

This comparison is motivated by prior work showing that the form of a demonstration affects both usability and the quality of the learned robot behavior. Demonstrations may be represented as keyframes, trajectories, or movement models, and these choices influence replay, refinement, and generalization~\citep{akgun2012keyframe,pastor2009learning,calinon2007incremental}. Recent work has also compared kinesthetic teaching with gamepad teleoperation for collaborative robot programming, showing that teleoperation can improve some demonstration-quality measures while requiring longer programming time~\citep{dallalba2025gamepad}. Gesture interfaces have been studied both for direct robot teleoperation and for communicating higher-level human intent, including multi-type gesture sentences that combine action gestures with task parameters~\citep{zhang2019gesture,deboer2023augmented,du2012markerless,vanc2023communicating}. In contrast to these works, our focus is a direct user-centered comparison of kinesthetic guidance, joystick teleoperation, and hand gestures on the same physical manipulation tasks, combining replay success, demonstration duration, user workload, user preference, and observed teaching errors.

Our results confirm that kinesthetic teaching is generally the easiest and most efficient modality, especially for tasks requiring precise alignment. However, the comparison also shows that it is not universally superior: joystick teleoperation performs very well on simple top-down grasping, and hand-gesture teaching performs better than expected. In several cases, gestures provide a viable contact-free alternative when direct kinesthetic teaching is unavailable, unsafe, or physically demanding.



\section{User Study Design}

\subsection{Robot Platform and Tasks}

We analyze a dataset of robot demonstrations collected in a user study on the Robothon Electronic Task Board~\cite{robothon}, see Fig.~\ref{fig:intro}. The experimental setup used a Franka Robotics Panda robot equipped with a RealSense D455 camera mounted in an eye-in-hand configuration tracking the manipulation process. 
The Robothon taskboard provides a compact benchmark environment for manipulation tasks involving grasping, tool use, contact, and cable handling.

We considered three manipulation tasks: \emph{Peg pick}, \emph{Probe measure}, and \emph{Cable wrap}. For task details, see Tab.~\ref{tab:task_scenarios}. Each task was presented in multiple scene variants, requiring the user to adapt the demonstrated behavior to the current task-board configuration. For example, the peg could be placed at different locations, the probe measurement point could be accessible or blocked by a door, and the cable could start from a different configuration. These variants make the dataset suitable for comparing teaching modalities across both simple demonstrations and demonstrations requiring adaptation to changed scene conditions.
Each demonstration recorded the robot pose, gripper state, and visual observations (camera) during task execution.

\begin{table*}[t]
\centering
\small
\begin{tabular}{p{0.08\linewidth} p{0.2\linewidth} p{0.28\linewidth} p{0.22\linewidth} p{0.05\linewidth}}
\toprule
\textbf{Scenario} & \textbf{Demonstrated behavior} & \textbf{Scene variants} & \textbf{Success criterion} & \textbf{\# Var.} \\
\midrule
\emph{Peg pick} &
Pick a peg and place it into a bowl. &
Peg placed at different task-board locations. &
Peg is successfully dropped into the bowl. &
2 \\
\midrule
\emph{Probe measure} &
Pick the probe, touch the measurement point, and place the probe into a bowl. &
Probe location and accessibility of the measurement point varied; in one variant, the measurement point was blocked by a door. &
Probe touches the measurement point and is then dropped into the bowl. &
3 \\
\midrule
\emph{Cable wrap} &
Wrap the cable around the task-board fixture. &
Initial cable configuration and starting side varied. &
At least two complete loops of cable are wrapped. &
2 \\
\bottomrule
\end{tabular}
\caption{Manipulation scenarios used in the user study. Each task was presented in multiple scene variants to elicit adapted demonstrations from users.}
\label{tab:task_scenarios}
\end{table*}

\subsection{Teaching Modalities}

As mentioned in Sec.~\ref{sec:intro}, we compare three teaching modalities for providing robot demonstrations: kinesthetic guidance, joystick teleoperation, and hand gestures measured by Leap motion sensor \cite{Weichert_Bachmann_Rudak_Fisseler_2013}. All teaching modalities were mapped to a common command representation consisting of robot motion control and gripper control, allowing the recorded demonstrations to be compared independently of the input device, see Tab.~\ref{tab:modalities}.

\begin{table}[t]
\centering
\begin{tabular}{p{0.18\linewidth} p{0.5\linewidth} p{0.22\linewidth}}
\toprule
\textbf{Modality} & \textbf{Motion control} & \textbf{Gripper control} \\
\midrule
Kinesthetic & Manual guidance with gravity compensation & Button \\
Joystick & Gamepad thumbsticks/buttons for translation and orientation & Buttons \\
Hand gestures & 6-DoF hand teleoperation & Hand open/close \\
\bottomrule
\end{tabular}
\caption{Teaching modalities used for robot demonstrations.}
\label{tab:modalities}
\end{table}

Kinesthetic teaching allows the user to physically guide the robot arm while gravity compensation is enabled. This gives direct control over the robot motion and can support precise end-effector placement, but it requires physical access to the robot. Joystick teleoperation avoids direct contact with the robot, but the user must decompose the desired motion into sequential translation and orientation commands. Hand gestures provide contact-free 6-DoF teleoperation and are therefore useful when direct physical guidance is impractical, although they may be affected by hand-tracking reliability and fine orientation control. 

Our hand teleoperation method is \textit{Clutch-based}: The user closes their hand to engage control of the robot end-effector. While engaged, the robot follows the user's 6-DoF hand motion. Opening the hand disengages control, so the user can reposition their hand without moving the robot. This interaction works similarly to a clutch-based 3D mouse or a drag-and-drop control in space.
Additional gestures are used for end-effector orientation. A circular finger motion controls rotation around the end-effector roll axis, while sideways rotation angle of the hand controls the pitch angle. Gesture recognition is based on the methods described in~\cite{vanc2023communicating} and implemented using the Teleoperation Gesture Toolbox\footnote{\href{https://github.com/imitrob/teleop_gesture_toolbox}{github.com/imitrob/teleop\_gesture\_toolbox}}.
See the paper website \footnote{\href{http://imitrob.ciirc.cvut.cz/publications/howtoteachrobot}{imitrob.ciirc.cvut.cz/publications/howtoteachrobot}} for more details and video examples of individual teaching trials using different modalities.

\subsection{Dataset and User Study}

Our user study involved eight participants aged 24--40. Seven participants had no prior experience with gesture teleoperation, and five had no prior experience with kinesthetic teaching. Before recording demonstrations, participants were introduced to the task setup and teaching modalities\footnote{\label{materials}See the complete user study narration material on our website \href{imitrob.ciirc.cvut.cz/publications/howtoteachrobot}{imitrob.ciirc.cvut.cz/publications/howtoteachrobot}}. 

We reuse the demonstration dataset from See \& Switch~\citep{vanc2026switchvisionbasedbranchinginteractive}, but analyze it here from the perspective of teaching modality rather than task branching.
During data collection, participants demonstrated the manipulation tasks using each modality. For each task, the environment was presented in multiple scene variants. When the scene configuration changed so that the previous demonstration was no longer appropriate, participants provided an adapted demonstration for the new condition, such as a peg located initially in a different location.

\paragraph{User Study Procedure}

Each participant was first introduced to the system terminology and task setup. To build intuition for the demonstration workflow for our single-arm manipulator, we asked participants to perform the selected tasks single-handedly before actually teaching the robot.
We also provided a small set of practical guidelines to improve safety and consistency during demonstrations, such as avoiding contact with the camera mount and keeping the cable in continuous contact with the task-board fixtures during cable wrapping.

\paragraph{Collected dataset}

Across 8 users and 3 modalities, the two 2-variant tasks and one 3-variant task yielded
$8 \times 3 \times (2 + 2 + 3) = 168$ recorded demonstrations.
Participants typically recorded one demonstration per task, modality, and variant. In approximately 3\% of cases, participants were allowed to repeat a demonstration and the previous attempt was discarded. Each demonstration variant was replayed and evaluated at least three times, yielding at least $3$ roll-outs. Approximately 4\% of demonstrations were considered unsafe (i.e., execution could damage taskboard or robot) and therefore contain only a single recorded roll-out; these cases are reported as unsuccessful in the replay-success analysis. In total, the dataset contains more than $600$ execution roll-outs.

\section{Evaluation Measures}

We evaluate the teaching modalities from both robot-performance and user-experience perspectives. For each modality, we analyze whether the demonstrated behavior could be replayed successfully, how much time the user needed to produce the demonstration, how demanding the modality was perceived to be, and what types of errors occurred during teaching or replay.

\subsection{Task Replay Success}

Replay success measures whether a recorded demonstration produced the intended task outcome when executed by the robot. Each recorded task variant was treated as a separate demonstration instance, and success rates were computed over the corresponding replay roll-outs. A replay was considered successful when the robot achieved the task-specific goal. The goals are described in Tab.~\ref{tab:task_scenarios}.

\subsection{Teaching Effort}

We use the demonstration duration as an objective proxy for teaching effort. The duration captures how long the user needed to produce a complete demonstration for a given task variant and modality. Longer demonstrations typically indicate slower motion, additional corrections, hesitation, or extra time needed for end-effector alignment. This measure is not intended to replace subjective workload, but it provides an observable indicator of how efficiently users could express the desired robot behavior.

\subsection{Subjective Workload and Preferences}

Subjective workload was measured using a modified NASA-TLX questionnaire after the demonstrations. Participants rated mental demand, physical demand, effort, frustration, and satisfaction with the achieved teaching quality, each on a 1--7 scale. For mental demand, physical demand, effort, and frustration, lower values are better (i.e., they indicate lower workload). For Performance satisfaction with performance, higher values indicate better perceived teaching quality. We also compute a composite score from mental demand, physical demand, effort, and frustration, and inverted performance satisfaction, calculated as $8 - \mathrm{performance~satisfaction}$.

In addition to workload, participants rated the overall difficulty of each task and answered preference questions about the teaching modalities. These included whether they would use each method again, how much they trusted the learned robot behavior, which method felt easiest, which method felt most precise, and which method they would prefer for similar tasks in the future.

\subsection{Observed Teaching Errors}

\begin{figure}
    \centering
    \includegraphics[width=0.99\linewidth]{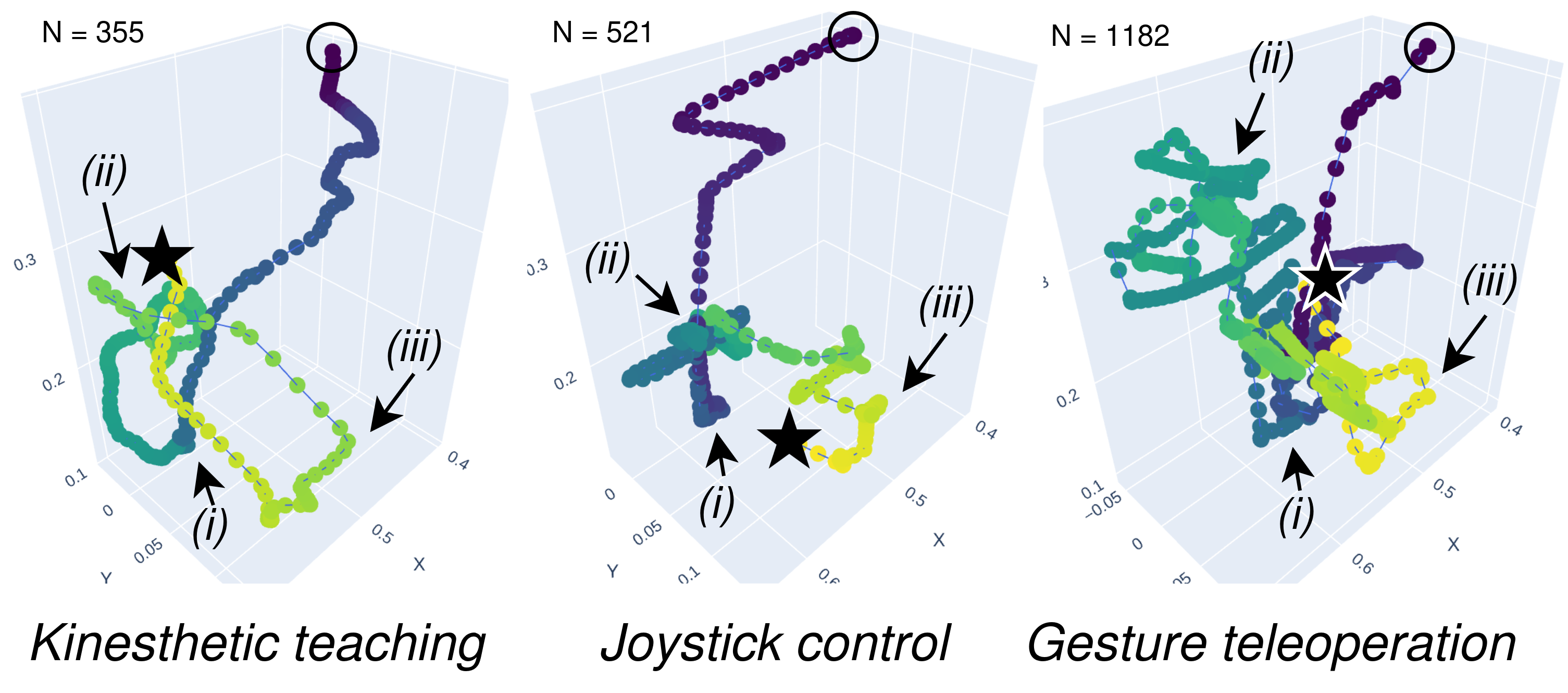}
    \caption{Example demonstration trajectories for \textit{Probe measure} task. ${\color{black} \circ}$ is demo start. ${\color{black} \star}$~is demo end. \emph{(i)} is probe pick, \emph{(ii)} is measurement with probe, and \emph{(iii)} is probe drop into a bowl. You can see that gesture probe measurement \emph{(ii)} is messy when the user tries to reorient the gripper and focus on target precision. $N$ is the number of time steps. Keypoint colors indicate the time encoding.}
    \label{fig:trajcharacteristics}
\end{figure}

We further analyzed common errors observed during demonstrations and replay. These included failed grasps, imprecise probe alignment, missing final placement actions, incomplete cable wrapping, unnecessary corrections, unstable motion, excessive orientation adjustments, and loss of hand tracking during gesture control. We also inspected trajectory characteristics across teaching modalities, focusing on whether demonstrations were smooth and direct, segmented into incremental motions, or noisy during fine end-effector orientation, see example differences in Fig.~\ref{fig:trajcharacteristics}.

\section{Results}

Tab.~\ref{tab:userstudy} summarizes replay success and average demonstration length for each task and modality. The easiest task was \emph{Peg pick}, where all teaching modalities achieved high replay success. This result is consistent with the task structure: picking a peg mainly required top-down grasping and only limited wrist reorientation.
Joystick teleoperation achieved the highest success on this task ($99.2\%$), followed by hand gestures ($92.7\%$) and kinesthetic teaching ($88.0\%$). 

The more orientation-sensitive tasks were harder. In \emph{Probe measure}, the user had to align the probe precisely with the measurement point, and kinesthetic teaching achieved the highest replay success ($83.0\%$). In \emph{Cable wrap}, the robot had to maintain a longer contact-rich motion around the task-board fixture. This was the most difficult task across all teaching modalities, with success rates between $48.8\%$ and $58.1\%$. Kinesthetic teaching again achieved the highest success on this task, but the overall lower values indicate that cable wrapping was difficult to demonstrate reliably for novice users.

\begin{table}[t]
    \centering
    \resizebox{\columnwidth}{!}{%
    \begin{tabular}{lccc}
        \toprule
        Modality & \textit{Peg Pick} & \textit{Probe measure} & \textit{Cable wrap} \\
        \midrule
        \multicolumn{4}{c}{\textit{Replay success (\%)}} \\
        \midrule
        \texttt{kin} & 88.0 & \textbf{83.0} & \textbf{58.1} \\
        \texttt{joy} & \textbf{99.2} & 80.3 & 53.1 \\
        \texttt{gst} & 92.7 & 72.2 & 48.8 \\
        \midrule
        \multicolumn{4}{c}{\textit{Demonstration length (s)}} \\
        \midrule 
        \texttt{kin} & \textbf{17.4 $\pm$ 4.2} & \textbf{19.5 $\pm$ 3.7} & \textbf{24.0 $\pm$ 5.4} \\
        \texttt{joy} & 26.1 $\pm$ 9.2 & 39.2 $\pm$ 18.6 & 65.3 $\pm$ 26.2 \\ 
        \texttt{gst} & 38.2 $\pm$ 28.7 &  41.6 $\pm$ 13.0 & 74.8 $\pm$ 38.1  \\
        \bottomrule
    \end{tabular}   
    }
    \caption{\textbf{User study results.} Replay success rate over at least three trials and average demonstration length for each modality and task.}
    \label{tab:userstudy}
\end{table}
\begin{figure*}[!ht]
    \centering
    \includegraphics[width=0.99\linewidth]{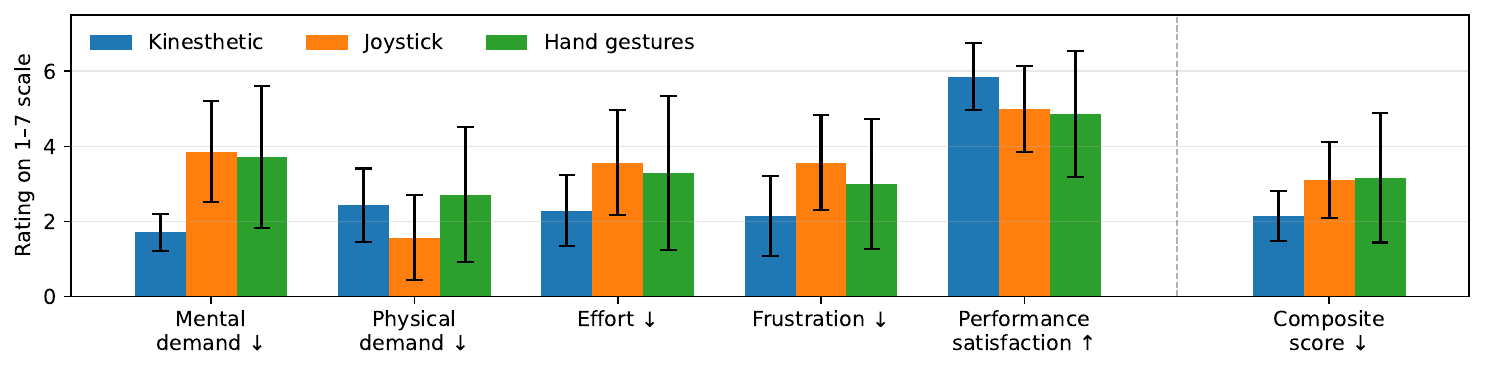}
    \caption{\textbf{Modified NASA-TLX workload ratings by modality.} Bars show mean ratings and error bars show standard deviations on a 1--7 scale. Lower values indicate better results for mental demand, physical demand, effort, frustration, and copmosite score, while higher values indicate better results for performance satisfaction. Composite score is computed from mental demand, physical demand, effort, frustration, and inverted performance satisfaction, where inverted performance satisfaction is calculated as $8 - \mathrm{performance~satisfaction}$.}
    \label{tab:nasa_workload}
\end{figure*}

Demonstration duration shows a clear difference in teaching effort. Kinesthetic teaching was consistently the fastest modality, with average durations between $19.5$ and $24.2$ seconds. Joystick and gesture demonstrations became substantially longer for tasks requiring fine orientation control or continuous manipulation. For \emph{Cable wrap}, joystick demonstrations lasted $61.4$ seconds on average, while gesture demonstrations lasted $73.5$ seconds. This suggests that direct physical guidance allowed users to produce task trajectories more efficiently, whereas joystick and gesture control required more corrections and slower alignment.

The trajectory examples in Fig. 2 illustrate these modality differences for Probe measure, where users had to reorient the gripper precisely while approaching the measurement point.
Joystick trajectories were more segmented, reflecting the incremental nature of the control input. Gesture trajectories were more variable, especially during fine end-effector orientation. This was most visible in \emph{Probe measure}, where users had to reorient the gripper precisely while approaching the measurement point.

The most common replay failures were manipulation errors rather than misunderstandings of the task goal. Typical failures included slip grasps, imprecise probe alignment, missing final placement actions, and incomplete cable wrapping. Gesture demonstrations were additionally affected by unstable hand tracking or noisy motion during fine orientation control. Joystick demonstrations were often limited by the need to decompose motion into separate translation and rotation commands, such as when users had to think about which button to press.

Fig.~\ref{tab:nasa_workload} reports the modified NASA-TLX questionnaire results. Kinesthetic teaching received the lowest mental demand ($1.71 \pm 0.49$), effort ($2.29 \pm 0.95$), and frustration ($2.14 \pm 1.07$), and the highest satisfaction with the achieved teaching quality ($5.86 \pm 0.90$). Joystick teleoperation had the lowest physical demand ($1.57 \pm 1.13$), but was rated as more mentally demanding ($3.86 \pm 1.35$) and more frustrating ($3.57 \pm 1.27$) than kinesthetic teaching. Hand-gesture teaching showed a similar overall workload to joystick teleoperation, with mental demand of ($3.71 \pm 1.89$), effort of ($3.29 \pm 2.06$), and frustration of ($3.00 \pm 1.73$), but also showed higher variability between participants.

Task-difficulty ratings followed the replay results. \emph{Peg pick} was perceived as the easiest task ($1.71 \pm 0.76$), \emph{Probe measure} was rated as moderately difficult ($2.71 \pm 1.38$), and \emph{Cable wrap} was rated as the most difficult task ($4.00 \pm 2.00$). User preferences also favored kinesthetic teaching: all participants selected it as the easiest method, five participants selected it as their preferred method for future use, and four respondents rated it as the most precise method. However, two participants preferred hand-gesture teleoperation overall, indicating that contact-free teaching was attractive for some users despite its higher variability.

Open questionnaire responses were consistent with these findings. Participants described kinesthetic teaching as intuitive and capable of expressing all robot motions. Joystick control was described as useful for large motions, but less convenient for precise orientation. Hand gestures were described as more natural than the joystick for some final adjustments, but users also reported hand-tracking failures, missing rotation control around the end-effector, and unintended vertical motion. These increased both the demonstration length and the user's frustration.

Finally, successful demonstrators produced smoother, more consistent trajectories, avoided unnecessary corrections, and appeared to plan the complete task before moving the robot. Demonstrations with excessive hesitation, repeated reorientation, or missing final actions were less likely to replay successfully. When users repeated demonstrations, later attempts were typically more direct and more reliable \cite{aliasghari2024non}.

 
\section{Discussion}

The results show that kinesthetic teaching remains the strongest baseline for efficient demonstration acquisition when direct physical interaction is possible. It produced the shortest demonstrations, the lowest subjective workload, and the highest perceived precision. This supports the expectation that physically guiding the robot is effective for novice users because it allows direct positioning of the end-effector without translating the intended motion into interface commands.
However, kinesthetic teaching is not universally preferable. It requires safe physical access to the robot, gravity compensation, and sufficient physical ability from the user. It also limits the robot's proprioceptive experience. Furthermore, it requires the user to be aware of the whole robot configuration, not only the end-effector pose. 
In practice, a user can guide the robot into a configuration that completes the immediate motion but is less suitable for replay or for the following part of the task. This issue was especially relevant in tasks involving contact or constrained motion, such as cable wrapping.

Joystick teleoperation had a different trade-off. It had the lowest physical demand and performed very well on \emph{Peg pick}, where the required motion was mostly translational and top-down. However, it became slower and more demanding for tasks requiring precise orientation, such as probe measurement and cable wrapping. This suggests that joystick control is suitable for simple or remote demonstrations, but less efficient when users must coordinate several translational and rotational degrees of freedom.
Hand-gesture teaching was the most interesting alternative. It was not the best modality overall, but it performed better than expected. Gesture teaching achieved high replay success on \emph{Peg pick} and remained usable for the more difficult tasks. Two participants even selected hand gestures as their preferred future method. This suggests that contact-free hand control can be intuitive for some users, especially when direct kinesthetic guidance is unavailable, unsafe, or physically demanding. Its main limitation is reliability: fine orientation control and hand-tracking errors can introduce noisy trajectories and user frustration.
The comparison also shows that the task strongly influences which modality is appropriate. For simple top-down grasping, joystick and gestures can be competitive with kinesthetic teaching. For tasks requiring precise end-effector alignment, kinesthetic teaching is more efficient. For long contact-rich manipulation, such as cable wrapping, all teaching modalities become difficult, and successful demonstrations depend not only on the interface but also on how well the user plans the complete robot motion.

Overall, the study supports a modality-dependent view of robot teaching from demonstrations. Kinesthetic teaching should be preferred when the robot is safely accessible and precision is required. Joystick teleoperation is useful when physical contact should be avoided and the task does not require complex continuous motion. Hand gestures provide a viable contact-free alternative, particularly for situations where kinesthetic teaching cannot be used, although improvements in tracking reliability and orientation control are needed.

The main limitation of this study is its small sample size of eight participants and its focus on a single robot platform and task board. The questionnaire results should therefore be interpreted as exploratory. In addition, only a small number of open-ended responses were collected, so qualitative comments are used mainly to support, rather than replace, the quantitative results. Future work should evaluate the same teaching modalities with more users, additional task types, and a more detailed analysis of how users improve over repeated demonstrations.

\noindent
{\bfseries\large Acknowledgement}\newline
Project was supported by the Czech Science Foundation (GA~ČR), Project No.~26-22610M, and by Robotics and Advanced Industrial Production (ROBOPROX), Reg.~No.~CZ.02.01.01/00/22\_008/0004590.

\vspace{3,16314mm}
\nocite{*}
\bibliographystyle{apalike_kuz}
\bibliography{references}

@misc{vanc2026switchvisionbasedbranchinginteractive,
      title={See and Switch: Vision-Based Branching for Interactive Robot-Skill Programming}, 
      author={Petr Vanc and Jan Kristof Behrens and Václav Hlaváč and Karla Stepanova},
      year={2026},
      eprint={2603.08057},
      archivePrefix={arXiv},
      primaryClass={cs.RO},
      url={https://arxiv.org/abs/2603.08057}, 
}

@article{Weichert_Bachmann_Rudak_Fisseler_2013, title={Analysis of the Accuracy and Robustness of the Leap Motion Controller}, volume={13}, rights={http://creativecommons.org/licenses/by/3.0/}, ISSN={1424-8220}, DOI={10.3390/s130506380}, abstractNote={The Leap Motion Controller is a new device for hand gesture controlled user interfaces with declared sub-millimeter accuracy. However, up to this point its capabilities in real environments have not been analyzed. Therefore, this paper presents a first study of a Leap Motion Controller. The main focus of attention is on the evaluation of the accuracy and repeatability. For an appropriate evaluation, a novel experimental setup was developed making use of an industrial robot with a reference pen allowing a position accuracy of 0.2 mm. Thereby, a deviation between a desired 3D position and the average measured positions below 0.2mmhas been obtained for static setups and of 1.2mmfor dynamic setups. Using the conclusion of this analysis can improve the development of applications for the Leap Motion controller in the field of Human-Computer Interaction.}, number={55}, journal={Sensors}, publisher={Multidisciplinary Digital Publishing Institute}, author={Weichert, Frank and Bachmann, Daniel and Rudak, Bartholomäus and Fisseler, Denis}, year={2013}, month={May}, pages={6380–6393}, language={en} }

@article{aliasghari2024non,
  title={How non-experts kinesthetically teach a robot over multiple sessions: diversity in teaching styles and effects on performance},
  author={Aliasghari, Pourya and Ghafurian, Moojan and Nehaniv, Chrystopher L and Dautenhahn, Kerstin},
  journal={Int. J. Soc. Robot.},
  volume={16},
  number={11},
  pages={2079--2105},
  year={2024},
  publisher={Springer}
}

@inproceedings{vanc2023communicating,
  title={Communicating human intent to a robotic companion by multi-type gesture sentences},
  author={Vanc, Petr and Behrens, Jan Kristof and Stepanova, Karla and Hlavac, Vaclav},
  booktitle={IEEE/RSJ IROS},
  pages={9839--9845},
  year={2023},
}

@article{akgun2012keyframe,
  title={Keyframe-based learning from demonstration: Method and evaluation},
  author={Akgun, Baris and Cakmak, Maya and Jiang, Karl and Thomaz, Andrea L.},
  journal={International Journal of Social Robotics},
  volume={4},
  pages={343--355},
  year={2012}
}

@inproceedings{pastor2009learning,
  title={Learning and generalization of motor skills by learning from demonstration},
  author={Pastor, Peter and Hoffmann, Heiko and Asfour, Tamim and Schaal, Stefan},
  booktitle={IEEE International Conference on Robotics and Automation},
  pages={763--768},
  year={2009}
}

@article{dallalba2025gamepad,
  title={Towards an intuitive industrial teaching interface for collaborative robots: gamepad teleoperation vs. kinesthetic teaching},
  author={Dall'Alba, Diego and Boriero, Federico},
  journal={The International Journal of Advanced Manufacturing Technology},
  year={2025}
}

@article{havoutis2019learning,
  title={Learning from demonstration for semi-autonomous teleoperation},
  author={Havoutis, Ioannis and Calinon, Sylvain},
  journal={Autonomous Robots},
  volume={43},
  pages={713--726},
  year={2019}
}

@article{zhang2019gesture,
  title={A gesture-based teleoperation system for compliant robot motion},
  author={Zhang, Wei and Cheng, Haobo and Zhao, Lijuan and Hao, Li and Tao, Ming and Xiang, Chen},
  journal={Applied Sciences},
  volume={9},
  number={24},
  pages={5290},
  year={2019}
}

@article{deboer2023augmented,
  title={Augmented reality-based telepresence in a robotic manipulation task: An experimental evaluation},
  author={de Boer, Thomas A. and de Winter, Joost C. F. and Eisma, Yke Bauke},
  journal={IET Collaborative Intelligent Manufacturing},
  volume={5},
  number={4},
  pages={e12085},
  year={2023}
}

@article{du2012markerless,
  title={Markerless Kinect-based hand tracking for robot teleoperation},
  author={Du, Guanglong and Zhang, Ping and Mai, Jianhua and Li, Zeling},
  journal={International Journal of Advanced Robotic Systems},
  volume={9},
  number={2},
  pages={36},
  year={2012}
}

@article{ravichandar2020recent,
  title={Recent advances in robot learning from demonstration},
  author={Ravichandar, Harish and Polydoros, Athanasios S and Chernova, Sonia and Billard, Aude},
  journal={Annual review of control, robotics, and autonomous systems},
  volume={3},
  number={1},
  pages={297--330},
  year={2020},
  publisher={Annual Reviews}
}

@ARTICLE{robothon,
  author={So, Peter and Sarabakha, Andriy and Wu, Fan and Culha, Utku and Abu-Dakka, Fares J. and Haddadin, Sami},
  journal={IEEE Robot. Autom. Mag.}, 
  title={Digital Robot Judge: Building a Task-Centric Performance Database of Real-World Manipulation With Electronic Task Boards}, 
  year={2024},
  volume={31},
  number={4},
  pages={32-44},
  doi={10.1109/MRA.2023.3336473}}

@article{argall2009survey,
  title={A survey of robot learning from demonstration},
  author={Argall, Brenna D. and Chernova, Sonia and Veloso, Manuela and Browning, Brett},
  journal={Robotics and Autonomous Systems},
  volume={57},
  number={5},
  pages={469--483},
  year={2009}
}

@incollection{billard2016learning,
  title={Learning from humans},
  author={Billard, Aude G. and Calinon, Sylvain and Dillmann, R{\"u}diger},
  booktitle={Springer Handbook of Robotics},
  pages={1995--2014},
  year={2016},
  publisher={Springer}
}

@inproceedings{calinon2007incremental,
  title={Incremental learning of gestures by imitation in a humanoid robot},
  author={Calinon, Sylvain and Billard, Aude},
  booktitle={Proceedings of the ACM/IEEE International Conference on Human-Robot Interaction},
  pages={255--262},
  year={2007}
}


\end{document}